\documentclass[runningheads]{llncs}
\usepackage[T1]{fontenc}
\usepackage{graphicx}
\usepackage{amsmath}
\usepackage{amsfonts}
\usepackage{amssymb}
\usepackage{dsfont}

\usepackage{nicefrac}
\usepackage{mathrsfs}
\usepackage{bm}
\usepackage{mathtools}

\newcommand{\bP}{\bm{P}}

\newcommand{\Ind}{{\mathds 1}}
\newcommand{\ind}[1]{\Ind_{\{#1\}}}

\newcommand{\restr}{\mathbf{\kern0.3ex
 \vert\kern-0.3ex}\backprime\kern0.3ex}

\DeclareMathOperator{\UD}{UD}
\DeclareMathOperator{\PD}{PD}

\newcommand{\bU}{\mathbf{U}}
\renewcommand{\bP}{\mathbf{P}}
\newcommand{\bPD}{\mathbf{PD}}

\usepackage{todonotes}

\usepackage{enumerate} 
\usepackage{xcolor}
\usepackage{tikz}
\usetikzlibrary{matrix,backgrounds}
\usepackage{cancel}

\begin{document}
\title{When fairness metrics fail: A utility-based perspective on $\varepsilon$-fairness}
%
%

\author{
    Tolulope Fadina\inst{1} and Thorsten Schmidt\inst{2}
}
\authorrunning{T. Fadina et al.}

\institute{
    \textsuperscript{} Actuarial Science and Risk Management, University of Illinois Urbana \\ Champaign, Illinois, USA 
    \email{tfadina@illinois.edu}
    \and Department of Mathematical Stochastics, University of Freiburg \\ Ermst-Zermelo Str. 2, 79104 Freiburg, Germany \email{thorsten.schmidt@stochastik.uni-freiburg.de}}

%
\maketitle              
\begin{abstract}

Fairness in decision-making processes is often quantified using probabilistic metrics. However, these metrics need not reflect the consequences of decisions for the affected individuals and groups. We develop a utility-based framework that incorporates these consequences into the assessment of fairness. Our main result shows that a decision-making process can satisfy
$\varepsilon$-fairness while nevertheless being maximally unfair once the utilities associated with its outcomes are taken into account. To address applications in which information on false negatives is unavailable, we also formulate a reduced setting that retains the essential elements of the utility-based fairness assessment.

We illustrate the framework through two applications: college admissions and credit-risk assessment. In both cases, probabilistic metrics may classify a decision-making process as approximately fair even though the corresponding utility outcomes are highly unequal. In the college-admissions example, our analysis shows that improving completion rates is necessary to achieve equality of utility across groups, while in the mortgage example, mitigating unfairness requires not only adjusting approval rates but also reducing the adverse consequences of default. These findings demonstrate that fairness assessments should account not only for the probabilities of different decisions but also for the consequences of those decisions.

\end{abstract}

\section{Introduction}
With the immense success of machine-learning algorithms, the aim of making fairer decisions,  particularly in fields where decisions have profound implications for individuals' lives, such as finance, healthcare, and employment, has also come within reach. 
Although numerous measures aimed at parity across different groups have been proposed, these often focus narrowly on statistical metrics.  
We argue that  focusing solely on narrow probabilistic metrics cannot capture the full picture of the question at hand, in particular the real-world consequences of  decisions maybe overlooked, see also \cite{selbst2019fairness,HeidariHoda2021}.

This article introduces a tractable utility-based approach that incorporates relevant aspects of the real-world context of the question at hand and therefore provides a more holistic understanding of fairness. This becomes particularly apparent when $\varepsilon$-fairness, introduced in  \cite{kearns2018preventing}, is considered: this concept is intended for situations in which exact parity is unattainable but small deviations from parity in the relevant probabilities are considered acceptable. We show that this expectation may fail once the decision-making process is evaluated in its real-world context. It  may even happen that a decision-making process satisfying  $\varepsilon$-fairness for a very small $\varepsilon>0$ is highly unfair in terms of the associated utilities, and we provide several examples illustrating this. In addition, we provide  sufficient conditions that can be applied, even when the context is known only approximately or uncertainty needs to be taken into account, to ensure the absence of disadvantages in terms of utility.

Since monetary measures are a special case of utility-based measures, our approach also allows us to conclude the absence of material inequality. 
We proceed by introducing a simple setup, which we call the \emph{standard setting} which includes the typically considered cases with four cases. Since in many practically relevant applications information on true and false negatives is not available, we also introduce a \emph{reduced setting}, in which these two cases are merged into one case. To indicate the broader scope of the utility-based approach, we also touch upon a \emph{general setting} in which hidden random variables determine the cases and correlation between default probability and utility is possible, paving the way for more complex treatments. Proofs are relegated to the appendix.

\subsection*{Related literature}
Regulation of AI is a highly important topic and we refer to \cite{Schmidt_Voeneky_2022,Giudici_2023} for recent accounts and further literature on this subject. Here, fairness is a core part of the regulatory process. We utilise the famous concept of describing human decisions via utilities, which is a well-established approach in economic contexts. The application to fairness dates at least back to \cite{arrow1973some}. Just recently, the treatment in the fairness context intensified, see \cite{Mitchell2021} for an overview. The works which are most related to our approach are \cite{heidari2019moral}, \cite{blandin2023generalizing}, \cite{DworkReingoldRothblum2023}, \cite{BenPorat2019ProtectingTP}, \cite{Hinsch_2022} and \cite{Wen2021algorithms}, see also \cite{RobertsonSchmidtHutterAwad2024}, \cite{GiudiciNeelakantanPavaranaSchmidt}. Credit modelling has of course been considered intensively, and we refer to \cite{MFE} for an overview. 
We mention that \cite{Liu2019} already pointed out that standard measures of fairness show difficulties in promoting the long-term being of the protected group in a feedback model. We highlight these difficulties from a different viewpoint; a feedback model is beyond the scope of this paper and left for future research. 
Fairness is also a major concern in the insurance literature, see e.g.\cite{lindholm2022discussion} and references therein. 
%


Since we also take uncertainty into account, we shortly refer  to \cite{gilboa1993updating,FadinaNeufeldSchmidt2019} for a detailed treatment and further references.
For a work that is intrinsically linked to fairness and uncertainty, see 
 \cite{DworkReingoldRothblum2023}. 

\section{The standard setting}

The fairness question we focus on concerns  $n$ individuals who are subject to a \emph{binary} classification by an algorithm. The underlying  state $Y_i$ of individual $i$ is a random variable, i.e.
$$Y_i \in \{0,1\}, \qquad 1 \le i \le n.$$
According to the usual convention in the machine learning literature, we associate the value $1$ with the ''good'' state, for example successfully completing a degree programme or repaying a loan.

Each individual $i$ is characterised by covariates, which we often do not observe. For the present analysis, we treat the covariates as fixed, meaning that we concentrate on individuals with similar covariates. A  detailed treatment taking these covariates into account is quite complex and therefore beyond the scope of this work. 

The fairness question arises from the presence of  a sensitive attribute. More precisely, each individual carries  a legally protected or ethically sensitive attribute 
$$ S_i\in \{0,\dots,N\}.$$

Of course, $Y$ is typically not known when the prediction is made - it is only available in labelled data or once the outcome has been realised.  
The algorithm assigns each individual $i$  a prediction $\hat Y_i$ of this outcome. We assume that 
$$\hat Y_i \in \{0,1\}$$ 
%
%
In the following, we will simply write $Y$ instead of $Y_i$ and $\hat Y$ instead of $\hat Y_i$ if there is no focus on a particular individual.

\subsection{The associated utilities}
The decision of the algorithm may affect an individual in very different ways, depending on the actual outcome $Y$. In the college admission setting the two possible decisions are: admitted, $\hat Y = 1,$ (predicting a successful completion of the degree programme)  or not admitted, $\hat Y =0$. In the first case, the utility will depend on whether the degree programme is successfully  completed (in this case $Y = 1$) or not (then $Y=0$), which we reflect by specific utilities assigned to each outcome. In the case where the applicant is not admitted, the utilities might also depend on $Y$: on the one hand, if the applicant is highly capable, she or he might also perform very well in another professional field. On the other hand, for an applicant who would not complete the degree programme, it may be preferable to avoid a substantial student-loan burden. It seems important to point out that typically  no data on $Y$ are available  for applicants who are not admitted -- which motivates the \emph{reduced setting} introduced later.

To capture the consequences associated with these different cases, we introduce the associated utilities. Utilities are a well-established tool for representing the preferences in rational decision-making\footnote{See for example Section 2.5 in \cite{FoellmerSchied}.}. 

\begin{figure}[t!]
\begin{center}
\includegraphics[width=5cm]{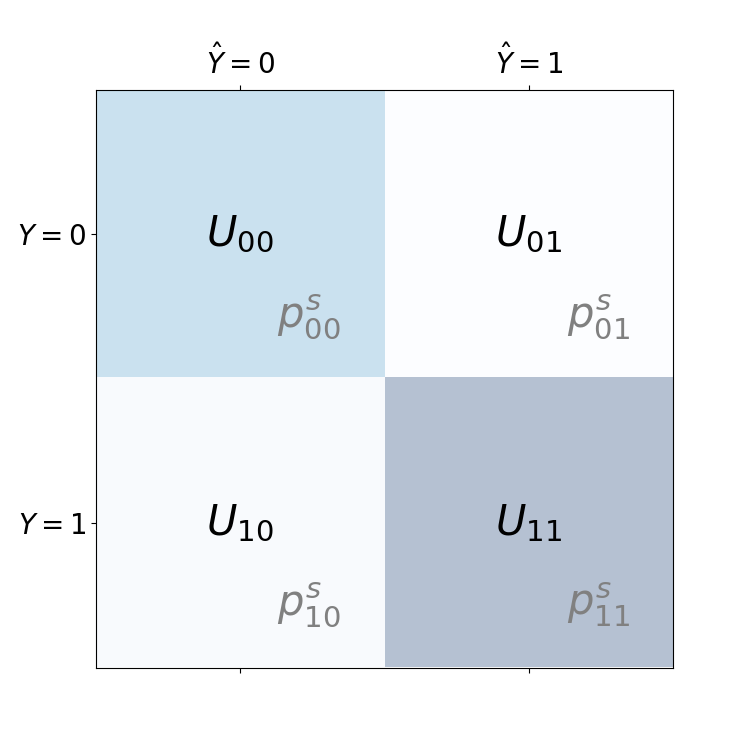}
\end{center}
\caption{Confusion matrix in our setting. For each group $s$ we associate with $Y=k$ and $\hat Y=j$ the utility $U_{kj}$ and the probability $p^s_{kj}$. The true positives are represented by $p^s_{11}$,  since $Y=1$ is considered the \emph{positive} case.}\label{fig1}
\end{figure}

Hence, to each  possible combination of $Y_i =k$  and $\hat Y_i=j$, we associate the utility 
$$ U_{kj}. $$
This means that, for a given outcome of the decision, say $\hat Y_i = 1$, the individual receives utility $U_{11}$ if $Y_i$ turns out to be equal to $1$, and $U_{01}$ if $Y_i$ turns out to be equal to $0$. Note that, in the standard setting, the utility does not depend on $i$, and is therefore the same for each individual, see Figure \ref{fig1}.

\begin{remark}[Individual utilities]
The utility parameters used here should be interpreted as common
benchmark valuations of the consequences associated with the possible outcomes. They provide a group-level assessment of fairness and should not be understood as asserting that every individual experiences a given outcome in the same way. Individual heterogeneity could be incorporated by allowing utilities to depend on individual characteristics and then averaging these utilities conditionally within each group and outcome category. Such an extension is not merely notational: it requires additional data and elicitation assumptions and, if the resulting conditional average utilities differ across groups, may require a reformulation of our results that rely on a common utility vector. We therefore restrict our attention to fixed benchmark utilities and interpret our conclusions as statements about fairness at the group level.
\end{remark}


\begin{example}[University admission]\label{ex:university admission}
 Admission to a university ($\hat Y=1$) offers a high utility if the degree programme is successfully completed. 
 Failing to complete the degree programme  may result in significant financial loss, including both direct costs and the foregone opportunities of alternative paths. Therefore, 
\begin{align} \label{U1}
U_{11} \gg U_{01}
\end{align}
indicating that the utility derived from successfully completing the degree programme, the case $(Y,\hat Y)=(1,1)$, is substantially greater (indicated by $\gg$) than that from not completing it  (the case $(0,1)$).
For the case of non-admission, we could neglect the  difference between $U_{10}$ and $U_{00}$. However,  it seems also plausible that an applicant with $Y=1$ (indicating the capability to successfully complete the degree programme) has better prospects  in other professional fields than an applicant with $Y=0$. This would suggest $U_{10}>U_{00}$.
\end{example}

\section{Fairness and inequality} 
It is common in the fairness literature to rely on a number of well-known statistical measures. We will refrain from a detailed discussion but  present only a simple example here. A more detailed discussion is subject to future work.

\subsection{Disadvantages in terms of utilities}
The central aim of our work is to measure inequalities in terms of  utilities. In contrast to many other approaches, we therefore do not rely on narrow statistical measures.
Note that if the utility function is chosen to be the identity, we consider  monetary outcomes directly,  and thus the detection of  \emph{material inequality} is included as a special case. However, working with utilities allows a more general viewpoint, which we adopt here.

To ease the exposition, we concentrate on two groups, i.e.\ $N=1$, while the arguments also extend  to the general case. 
It is typically considered unfair if the algorithm produces  unjustified disadvantages.
Disadvantage will be measured in terms of  the \emph{utility difference} between the groups. We do not aim to define a distance, but rather a directed quantity. To this end, we refer to $s=0$ as the \emph{standard group} and $s=1$ as the \emph{protected group}. 
We denote by $P_s$ the probability conditional on $S=s,$ for $s\in\{0,1\}$ and by $E_s$ the associated expectation.

Writing $U=U_{Y,\hat Y}$, a \emph{disadvantage} (for the protected group) is indicated by a positive utility difference $\UD$ where 
\begin{align}
    \UD := E_0[U] - E_1[U].
\end{align} 
If the utility difference is negative, the outcome is better for the protected group than for the standard group, such that there is no disadvantage. 

Small disadvantages might be acceptable, and we fix a threshold $\tau \ge 0$ for acceptable differences in terms of utility. A utility difference above $\tau$ is  classified as a non-negligible disadvantage for the protected group, which leads to the following definition\footnote{A similar notion with a threshold was already introduced in \cite{blandin2023generalizing}, although with less  focus on utility comparison than here. Quite similar in spirit, one could replace $\UD$ by its absolute value, $|\UD|$ and provide analogous results.}.
\begin{definition}
\label{def:disadvantage}
     A decision rule induces a disadvantage at level $\tau$, if
    \begin{align}
        \UD > \tau.
    \end{align}
\end{definition}
In the following,  by \emph{fairness} we refer to the case in which there is no disadvantage (for the protected group) at level $\tau=0$. Sometimes we also use the term \emph{fairness in utility} to emphasise the difference to other fairness metrics.

\subsection{Expected utility}

Our first result computes the expected utility  in our setting. It turns out that this expression can be simplified in the case with two groups only, while an analogous expression can obtained in a similar manner for arbitrarily many groups. 

\begin{proposition} \label{lem:EU}
Assume that there are only two groups, i.e.\ $N=1$. Then, 
    the expected utility for group $s\in\{0,1\}$ is given by
    \begin{align}
        E_s[U] &=  P_s(Y=1,\,  \hat Y=1)  \cdot ( U_{11} - U_{01} ) \notag \\
        &+P_s(Y=0, \, \hat Y=0) \cdot( U_{00} - U_{10}) \notag \\
        & +P_s(\hat Y=1) \cdot (U_{01}-U_{10}) +
     U_{10}.
  \label{eq:EUi 1}
    \end{align}
\end{proposition}
To provide an easy access to this formula we introduce the following notation:
\begin{align*}
    \bU := \begin{pmatrix}
     U_{11}-U_{01}\\ U_{00}-U_{10} \\  U_{01}-U_{10}\end{pmatrix}, \qquad 
    \bP_s := \begin{pmatrix}
        P_s(Y=1, \hat Y=1) \\
        P_s(Y=0, \hat Y=0) \\
        P_s(\hat Y = 1),
    \end{pmatrix}
\end{align*}
such that
\begin{align}\label{eq:EsU}
    E_s[U] = \bU \cdot \bP_s+U_{10}. 
\end{align}

The following proposition draws the connection to probabilities: if the \textbf{joint} probabilities of both groups coincide, then fairness holds. The representation above also implies that all partial statistics (conditional use accuracy, positive predictive value, etc.) \textbf{do not guarantee fairness} without further assumptions.
\begin{proposition}[Equal joint probabilities imply fairness]
 \label{prop: 3.6}
Consider the case with two groups, i.e. $N=1$.
If 
\begin{align*}
P_0(\hat Y = i  \,,\,  Y = j) = 
P_{1}(\hat Y = i  \,,\,  Y = j), 
\end{align*}
for at least three pairs $(i,j) \in \{0,1\}\times \{0,1\}$,
then 
the decision rule does not induce a disadvantage.
\end{proposition}
Since probabilities sum up to one, we do not need all four pairs $(i,j)$, and can reduce to three in the above results. 

\begin{remark}[Connection to risk-measures]
The  application of risk-measures in fairness question has recently been proposed in \cite{williamson2019fairness,paes2024multi,mehta2023stochastic} At a first glance, it is tempting to expect equivalence between utility-based approaches and risk-measure approaches, due to the well-known equivalence of the approaches, see for example Chapter 4.9 in \cite{FoellmerSchied} . A deeper look reveals that our approach works on utility differences, while the risk-measure approaches work on the random variable conditional on the group $S$. The latter is therefore not able to control explicitly the inequalities between groups (measured in differences) but rather measures an aggregated quantity which highly depends on the demographic distribution of the groups (through the distribution of $S$). From our current viewpoint, the situation is therefore very similar to the one which will be formulated in Proposition \ref{prop:5}: a bound on the fairness risk-measure will not imply that the utility difference is small.
\end{remark}

\subsection{Computing the utility difference}
We now introduce the appropriate tools to compute the utility difference $\UD$  in our setting with two groups, i.e.\ when $N=1$.
With the associated conditional probability measures $P_0$ and $P_1$ we introduce the difference
\begin{align}
    \PD(A) := P_0(A) - P_1(A)
\end{align}
for all measurable sets $A$. From equation \eqref{eq:EsU} we obtain that
\begin{align}\label{eq:UD with PD}
    \UD = \bU \cdot \bPD 
\end{align}
where the vector $\bPD=\bP_0-\bP_1$ is given in terms of  the probability differences by
$$ 
\bPD := \begin{pmatrix}
        \PD(Y=1, \hat Y=1) \\
        \PD(Y=0, \hat Y=0)\\
        \PD(\hat Y = 1)
    \end{pmatrix}.$$

\section{The unfairness of $\varepsilon$-fairness}
A general notion of fairness relies on a  measure which describes the fairness of a certain decision rule $\hat Y$. 
From the above discussion, it becomes clear that the question of fairness depends on the full joint distribution in the different groups. As a consequence, fairness can be related to a certain \emph{distance} between measures. Let us denote this distance by $d$. We say that the estimator $\hat Y$ achieves $d$-fairness if
\begin{align}\label{eq: d-fairness}
    d(P_{s_1},P_{s_2})=0.
\end{align}
In reality, it will typically not be possible to achieve a vanishing distance. In particular, since the probabilities need to be estimated, even if the distance vanishes on the estimated probabilities, this  does not guarantee
that the distance between the true underlying probability measures vanishes (in general  not even with high probability).
Therefore it is natural to relax this notion in the following way: $\hat Y$ achieves $(d,\varepsilon)$-fairness if
\begin{align}\label{eq: d,epsilon-fairness}
    d(P_{s_1},P_{s_2}) \le \varepsilon.
\end{align}
If the metric is clear from the context, we simply talk about $\varepsilon$-fairness. 

The main goal of this section is to show that under certain assumptions $(d,0)$-fairness ensures utility-fairness, while this is not the case for $(d,\varepsilon)$-fairness. In fact, we provide an example which is $(d,\varepsilon)$- fair for an $\varepsilon>0$ arbitrarily close to zero but is unfair in the sense we consider here, i.e.\ measured in terms of  utilities and, in fact, can exhibit an arbitrarily large utility difference.

For all usual metrics $d$, a positive distance
\begin{align} \label{eq:d}
d(P_0,P_1) >0
\end{align}
implies that at least one of the distances  $|\PD(Y=0, \, \hat Y=0)|$, or $|\PD(Y=1,\,  \hat Y=1)|$ is positive, hence
\begin{align} 
\label{eq:d2}
|\PD(Y=0, \, \hat Y=0)| + |\PD(Y=1,\,  \hat Y=1)| > 0.
\end{align}

The following results shows that $\varepsilon$-fairness can be maximally unfair. In particular, even if the distance of the probability measures is arbitrarily small but not zero, the utility difference can be arbitrarily large.
\begin{proposition}\label{prop:5}
Assume that 
\eqref{eq:d2} holds.
     Then,  for every $K > 0$, there exist parameters $U_{00},\dots,U_{11}$ such that
    $$ \UD \ge K. $$
\end{proposition}
We refer to the simple proof of this proposition in the appendix which illustrates that the utilities can be chosen so as to produce an arbitrarily large utility difference. In practice, however, there should be natural bounds on the utilities available. For this case we develop a number of tools in the following.

Without further assumptions, we can at most hope for the following upper bound. Having Equation \eqref{eq:UD with PD} in mind, we say that \emph{utility differences are bounded by $K$} if the vector $\bU=(U_1,U_2,U_3)^\top$ satisfies
$$ \max\{ | U_1|,|U_2|,|U_3| \} \le K. $$
Moreover, we say that  the \emph{probability differences are bounded by $\varepsilon$}, if the vector $\bPD=(\PD_1,\PD_2,\PD_3)^\top$ satisfies
$$ \max\{ | \PD_1|,|\PD_2|,|\PD_3| \} \le \varepsilon. $$

\newpage
\begin{proposition}\label{prop:rough}
Assume that utility differences are bounded by $K>0$ and probability differences by $\varepsilon >0 $. Then, there is no disadvantage larger than
$$ 3 \hspace{0.2ex} \varepsilon  K.$$
\end{proposition}

\subsection{Demographic parity and equalized odds}
There is an extensive discussion about which metric should be chosen to study fairness of an algorithm\footnote{See for example \cite{kusner2017counterfactual} for related definitions and further literature. }.
One metric which is of interest in the fairness discussion is \emph{demographic parity}. In our context, demographic parity arises naturally, see Equation \eqref{eq:UD with PD}. Indeed, demographic parity holds if
$$ P_0(\hat Y=1) = P_1(\hat Y=1),$$
and hence $\PD(\hat Y=1)=0.$ We say that \emph{base-rate parity} holds, if
$$ P_0(Y=1) = P_1(Y=1).$$

A further metric which is often discussed, is \emph{equalized odds}. This measure requires equal true positive and false positive rates, i.e.\
\begin{align}
    P_0(\hat Y=i|Y=i) = P_1(\hat Y=i|Y=i),\qquad  i=0,1.
\end{align}
The following proposition shows that  equalized odds together with base-rate parity implies fairness.
However, since base-rate parity cannot be controlled and will often fail to hold, this should rather rarely be applicable in practice. 
\begin{proposition}
    Assume that base-rate parity and  equalized odds hold and that $U_{01}=U_{10}$. Then, there is no disadvantage.
\end{proposition}
We note that demographic parity together with equalized odds is \emph{not} sufficient to conclude fairness.

\section{College admission}
\label{ex:win win}

As an example of our framework, we study college admission in further detail, extending Example \ref{ex:university admission}.
The study in \cite{webber2016college} shows that besides, age, education, race and gender also cognitive and non-cognitive abilities play an important role. For the first case we consider two groups which have median abilities but otherwise only differ in the protected attribute. 

For the numerical illustration below we use 
$$ U_{11}=170{,}000 $$
as an illustrative monetary calibration for the benefit associated with a successful college outcome. This is motivated by the study mentioned above, which is taken from the ``some college'' category therein.

\subsection{A first, rough estimate}
Proposition \ref{prop:rough} gives us a first rough estimate on possible disadvantages. 

Anticipating the student-loan specification \eqref{eq:U01} below, we obtain $K=230{,}000$ as minimal bound on the utilities in this case. 
If a small $\varepsilon$ could be achieved this would already lead to a good bound. 
In the example below we consider a difference in conditional completion probabilities of $\delta=20\%$. For $q_0=80\%$, this gives a probability difference of $q_0\delta=0.16$ 
%
and so
$$ \UD \le 0.16 \cdot 230{,}000 = 36{,}800$$
which is substantial. Despite the fact that  demographic parity holds in this case as we detail in the following.

\subsection{The simplest case}
Additionally, we assume $U_{01}=0$, since we see this as our reference case here. Also let us assume for this case that $U_{10}=U_{00}=0$. It is clear that university admission is highly attractive  and everybody should have equal opportunities to participate. 

We again assume that group $0$ is admitted with probability $q_0\in [0,1]$ and the protected group $1$ with probability $q_1 \in [0,1]$.
The algorithm has some forecasting abilities, and hence $Y$ is not independent from $\hat Y$. However failure rates are still reported to be high\footnote{\cite{webber2016college} reports up to 40\%.} and we assume that the forecast is little worse in the protected group, possibly due to lack of data or other  factors. We introduce $\delta>0$ as follows: 
\begin{align*}
     P_0(Y=1 \,| \,\hat Y=1) &=  P_1(Y=1 \,|\, \hat Y=1) + \delta \\
  &> P_1(Y=1 \,|\, \hat Y=1) =:q_1(1,1);
\end{align*}
where we denote  $q_s(i,j)=P_s(Y=i \,|\, \hat Y=j)$. Then,
    \begin{align}
    \UD &= \bU \cdot \bPD \notag\\
    &= \begin{pmatrix}
        U_{11} \\0 \\ 0
    \end{pmatrix} \cdot 
    \begin{pmatrix}
        q_0 \cdot (q_1(1,1)+\delta) - q_1 \cdot q_1(1,1) \\
        (1-q_0) \cdot q_0(0,0) - (1-q_1)\cdot q_1(0,0) \\
        q_0-q_1 
    \end{pmatrix} \notag \\
    &= U_{11} \cdot \big( q_1(1,1) \cdot (q_0-q_1) + q_0 \cdot \delta\big).
\end{align}
The simplicity of this formula stems from our assumptions, since only $U_{11}$ was assumed to be non-vanishing, while all other utilities vanish.

At this point two things become clear: First,  under demographic parity, (i.e.\ $q_0=q_1$), 
$$ \UD_{q_0=q_1} = U_{11} \cdot q_0 \cdot \delta . $$
Hence, for $q>0$, demographic parity eliminates the disadvantage only if $\delta=0$. If $q=80\%$ and $\delta=20\%$, the utility difference equals $27{,}200$.

Second, we compute the smallest level of admission probabilities for the protected group which implies no disadvantages:
 assume that $q_1(1,1)=70\%$, i.e.\ 30\% failure rate. Then, 
$$ q_1^* = \frac{   q_1(1,1) \cdot q_0 + q_0 \cdot \delta}{  q_1(1,1)} = q_0 \cdot \Big( 1+ \frac{\delta}{ q_1(1,1)} \Big) \approx 1.0286. $$
 Thus $q_1^*>1$, showing that no feasible admission rate for the protected group can eliminate the disadvantage through admission probabilities alone. Even if all members of the protected group are admitted ($q_1=1$), a utility disadvantage of $3{,}400$ remains. Hence, achieving no disadvantage additionally requires reducing the difference in completion rates represented by $\delta$.

It turns out that more needs to be done to achieve fairness: the failure rates of the protected group need to be reduced, a surprisingly strong result from our simple framework.  

\begin{figure}[t]
    \begin{center}
        \includegraphics[width=3.9cm]{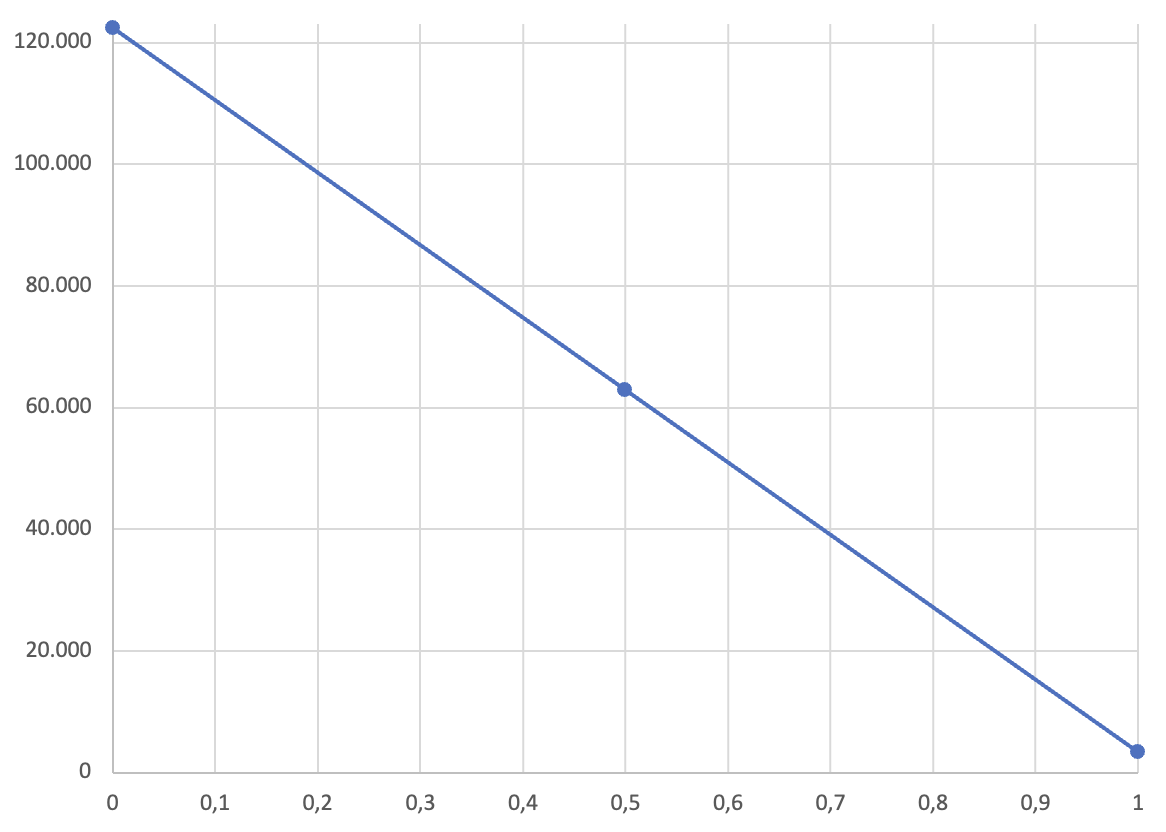}\ \includegraphics[width=3.9cm]{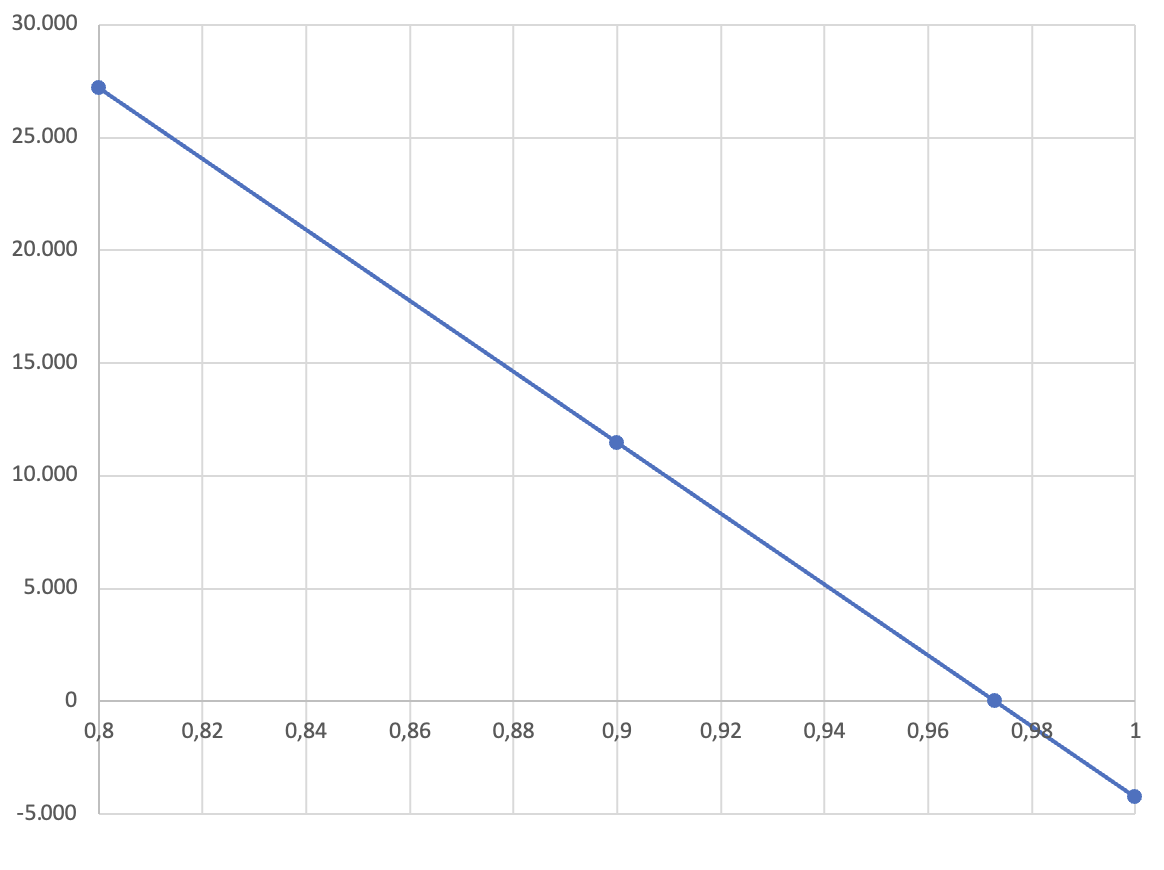}
    \end{center}
    \caption{Utility differences for the college admission. \underline{Left:} The utility difference as function of $q_1$. Even if the admission rate of the protected group equals $q_1=100\%$, there remains a disadvantage of ca. 3,400. \underline{Right:} The  case with a fixed number of admitted persons. We admit 1000 persons out of 1250, where 250 belong to the protected group. Starting from an equal admission rate of 80\% we allow additional persons from the protected group, meanwhile diminishing the number of admissions from the other group. In this case one is able to achieve equal utilities by admitting approx. 97,3\% of the protected group. }\label{fig:university admission}
\end{figure}

We illustrate the situation in Figure \ref{fig:university admission}. On the right hand side we modify the setup slightly, by considering a fixed number of possible admissions, which improves the situation a bit (we do not detail the simple calculations).

\subsection{Incorporating student loans}


We now revisit this case and incorporate student debt when studies are not completed successfully. As an illustrative high-debt scenario, motivated by the magnitudes considered in \cite{webber2016college}, we set
\begin{align}\label{eq:U01}
    U_{01} = -60{,}000.
\end{align}
In this case, we obtain 
\begin{align}
    \lefteqn{\UD = \bU \cdot \bPD} \notag\\
    &= \begin{pmatrix}
        U_{11} -U_{01}\\0 \\ U_{01}
    \end{pmatrix} \cdot 
    \begin{pmatrix}
        q_0 \cdot (q_1(1,1)+\delta) - q_1 \cdot q_1(1,1) \\
        (1-q_0) \cdot q_0(0,0) - (1-q_1)\cdot q_1(0,0) \\
        q_0-q_1 
    \end{pmatrix} \notag \\
    &= (U_{11}-U_{01}) \cdot \big( q_1(1,1) \cdot (q_0-q_1) + q_0 \cdot \delta\big) \notag\\
 &+U_{01}\cdot (q_0-q_1).
 \label{eq15}
\end{align}
We first observe that now $U_{11}$ is replaced by
$$ U_{11}-U_{01}=230{,}000,$$
such that the previously discussed effect is increased by  approximately $35\%$. 
Demographic parity ($q_0=q_1$) in this case eliminates the second component of the utility difference (see Equation \eqref{eq15}).

\subsection{Taking wealth into account}
Up to now we presented fairly general results and concentrated on monetary inequality. However, the utility-based framework offers much more, for example different wealth between the groups or even on an individual level can be taken into account through a utility function. In this case, the utilities will depend on the group. This can be incorporated into our framework, but is beyond the scope of this paper and therefore left for future research.

\section{Sufficient conditions}
The preceding results showed that one has to be extremely careful when considering fairness, in particular when the context is left aside. In the following we will show how to come up with useful estimates which allow to conclude a suitable level of fairness even when the utilities are not known precisely. In the case when some utilities vanish, we may reduce our requirements even further and can conclude fairness on the basis of some well-known requirements.

\subsection{Conditional use accuracy}
\label{sec:CUA}
While many conditional quantities like true positive rate or equalized odds typically do not imply equality of the unconditional probabilities they are  difficult to connect to Equation \eqref{eq:EsU}.  

This is simpler with the conditional use accuracy, which we will exploit now. In principle this can also be applied to equalized odds, but not without a few additional steps, such that we do not consider this quantity here.

According to \cite{caton2024fairness} Section 3.2.2, equality in the \emph{conditional use accuracy} (CUA) holds, if 
\begin{align*}
    \PD(Y=i\,|\, \hat Y=i) = 0
\end{align*}
for $i=0,1$.

Recall that by \emph{no disadvantage} in the sense of Definition \ref{def:disadvantage} we mean a vanishing (or negative) utility difference, i.e.\ $\UD \le 0$.
For the following proposition we denote by $\Gamma \in [0,1]$ the number
\begin{align}
    \Gamma = \max\{P_0( Y=1\,|\, \hat Y=1); 
    P_0( Y=0\,|\, \hat Y=0) \}.
\end{align}
Note that under CUA, $P_0$ can be replaced by $P_1$ for the computation of $\Gamma$. 
\begin{proposition}Assume that 
equality of the conditional use accuracy holds. 
\begin{enumerate}[(i)]
    \item If $P_0(\hat Y=1)=P_1(\hat Y=1)$, then there is no disadvantage.
    \item If  $|P_0(\hat Y=1)=P_1(\hat Y=1)|\le \varepsilon$ and the utility differences are bounded by $K>0$, then there is no disadvantage larger than
    $$  (1+2 \Gamma) \cdot  \varepsilon \cdot K.$$
\end{enumerate}\label{prop: 8}
\end{proposition}
Im comparison to Proposition \eqref{prop:rough}, CUA allows to reduce the general bound further, in particular if $\Gamma$ is small. 
%
The second result shows that if $  (1+2 \Gamma) \varepsilon K \le \tau$ for the level $\tau$ in Definition \ref{def:disadvantage}, there are only negligible disadvantages. In our examples $K$ turns out to be quite small, such that this seems to be a result highly relevant for practical cases.

\subsection{Independence}
An abstract criterion for fairness (on the probability level only) which immediately makes sense is independence, see \cite{caton2024fairness} Section 3.1, for example. We will show that independence also implies absence of disadvantages and, hence, fairness in the sense we consider here - but is not necessary. In essence, it is to strict. It also turns out that independence of the estimator $\hat Y$ and the group alone is \textbf{not} sufficient.

To consider independence, assume that the group $S$ is a random variable. Our probability measures $P_s$ are then obtained by conditioning on $S=s$. 

\begin{proposition}
Independence of $(Y, \hat Y)$ and $S$ implies absence of a disadvantage. 
\end{proposition}

\section{The reduced setting}
In many situations there is not enough information about the probability of the cases $\{Y=0, \hat Y=0\}$ and $\{Y=1,\hat Y=0\}$: in the college admission this would refer to the probability of successfully completing the studies while not being admitted to the university, a probability which is clearly difficult to estimate. 

\begin{figure}[t!]
\begin{center}
\includegraphics[width=5cm]{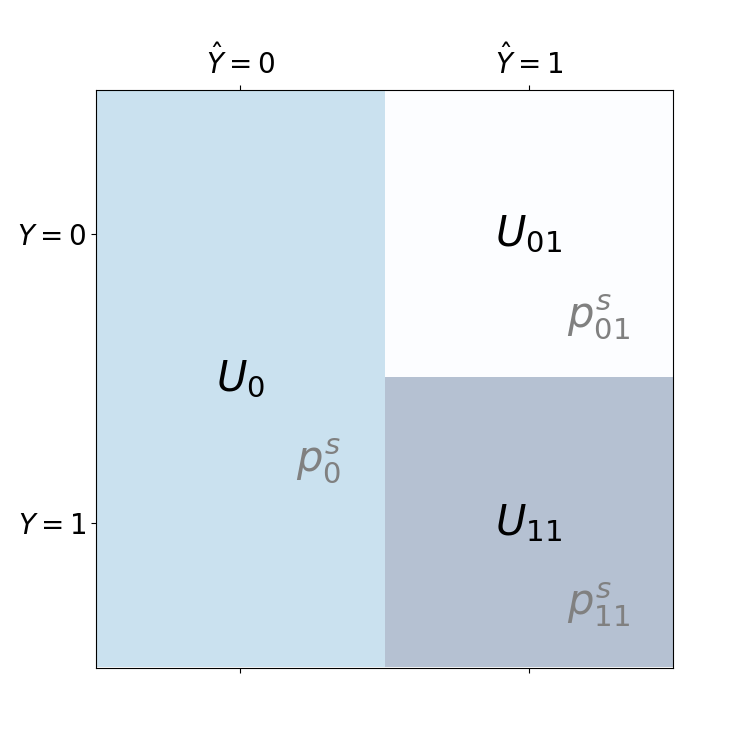}
\end{center}
\caption{Confusion matrix in the reduced setting. The cases 00 and 10 are now collapsed to the case $0$ (where $\hat Y=0$).}\label{fig2}
\end{figure}

To additionally capture these important situations within our framework, we introduce the \emph{reduced setting}: here, both cases (00 and 10) are collapsed into the case 0 (referring to $\hat Y=0$) and hence only the probability
$$ p^s_0 := P_s(\hat Y=0)$$
is needed, 
see Figure \ref{fig2} for an illustration.

Note that the results for the reduced setting can be obtained as a special case of the standard case by letting $U_{10}=U_{00}=U_0$.
\begin{lemma}
The expected utility in the reduced setting is
\begin{align}\label{eq:EsU2}
    E_s[\tilde U] = \tilde \bU \cdot \tilde \bP_s + U_0
\end{align}
for $s \in \{0,1\}$ 
with
\begin{align*}
    \tilde \bU := \begin{pmatrix}
     U_{11}-U_{01}\\ U_{01}-U_{0}\end{pmatrix}, \qquad 
    \tilde \bP_s := \begin{pmatrix}
        P_s(Y=1, \hat Y=1) \\
        P_s(\hat Y = 1).
    \end{pmatrix}
\end{align*}
\end{lemma}
In the college example we were already working in a reduced setting since we assumed $U_{00}=U_{10}=0$.

\subsection{Sufficient conditions in the reduced setting}

We adopt the conditional use accuracy to this setting, compare  Section \ref{sec:CUA}. 
Since $P_s(Y=0\mid\hat Y=0)$ is not available in the reduced setting, we replace the corresponding condition by equality of the unconditional rejection probabilities and call the resulting criterion (conditional) use accuracy, (C)UA.
It holds if
\begin{align}
\begin{aligned}
\PD(Y=1\,|\,\hat Y=1) &=0, \\
\PD(\hat Y=0) &=0.
\end{aligned} \label{(C)UA}
\end{align}
We immediately obtain the following result from Proposition \ref{prop: 8}
\begin{proposition}Assume that 
equality of the (conditional) use accuracy holds, i.e.~\eqref{(C)UA}. Then there is no disadvantage.
\end{proposition}

\section{A mortgage example}
As an example where it is important to consider the context for an in-depth analysis of fairness, we  consider credit ratings used in the context of mortgages. 
It also illustrates the importance of the reduced setting, since for rejected applicants we do not observe whether they would have defaulted had the mortgage been granted.
%
%
In contrast to smaller credits, the mortgage case is much more involved. We take some time to show how a fundamental model can be developed and what are the conclusions drawn. We start with a simplified setting and then illustrate the full power of the utility-based approach.

\begin{figure}[t!]
    \centering
    \includegraphics[width=6cm]{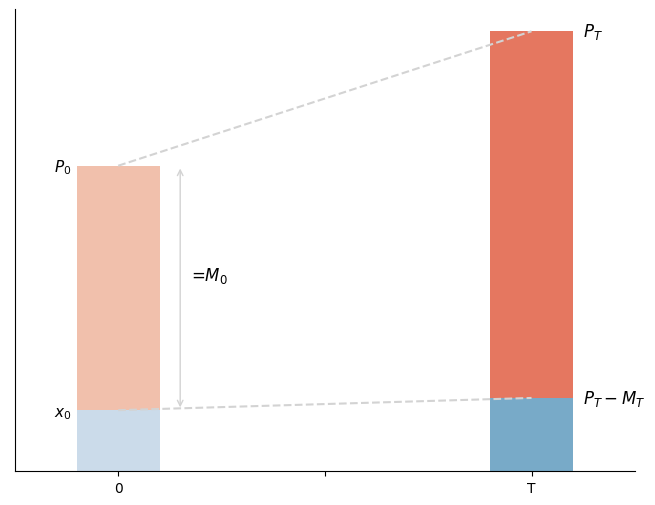}
    \caption{Illustration of the mortgage example}
    \label{fig:enter-label}
\end{figure}
For the mortgage, we look at the evolution over time, starting at time $t=0$ and ending with the payback time $T>0$. 
Consider a mortgage of size $M_0$ for a house of price $P_0$ and assume that the individual has liquid capital $x_0$. For simplicity, we assume $M_0=P_0-x_0$. The value of the house changes over time and its value at time $T$ is denoted by $P_T$. The borrower saved the amount $x_T$ until time $T$.

First, if $X_T \ge M_T$ there is no default. Second, if $X_T < M_T$, there is default and the house has to be sold\footnote{Here we focus on the case in Germany, where the house owner always is obliged to pay back the credit. This differs in other states, like in the U.S. for example.}. This typically can only be sold at a lower price, which we denote by $P_T \cdot (1-\lambda)$ with $\lambda \ge 0$.

Now, if $P_T\cdot (1-\lambda) + X_T \ge M_T$, the bank has no loss - the full mortgage can be covered. However, the borrower ends up with a possibly small fraction 
$$ P_T \cdot (1-\lambda) + X_T - M_T.$$
If $P_T\cdot(1-\lambda) + X_T < M_T$, the bank receives what is left,
$$ \big( P_T \cdot (1-\lambda) + X_T \big)^+,$$
and faces a (typically small) financial loss.

This already gives the impression that the risk for the bank is quite small, but the risk for the borrower is substantial. In particular in the case of default, the situation for the borrower is difficult and associated with a highly negative utility.

\subsection{Reduced setting}
We first simplify the setting and view the approach from a purely data-based point of view, as it is often done in this case. In this case we end up in the reduced setting introduced above.

We note that in the reduced setting we have to specify utilities for the cases $U_{11},\ U_{01}$ and the case $U_0$ together with associated probabilities. For simplicity we neglect discounting.
We think of a house costing $P_0=300{,}000$ and the borrower has capital of 30\%, such that the needed mortgage equals $M_0=210{,}000$. Assume that with a probability of  $95\%$ no default happens. The associated utility is
$$ U_{11} = 400{,}000-90{,}000 = 310{,}000$$
through the rise in the value of the house minus the initial capital. On the other side, in case of default the house can be sold at 300{,}000 in the good case and 200{,}000 in the bad case and we consider the average\footnote{For simplicity, we assume that the earnings over the 10 years are enough to approximately cover the $M_0=210,000$ repayment of the mortgage (and no default to the bank occurs). For the case of no-mortgage we include a little less, i.e.\ 200,000.}. Hence,
$$ U_{01} = 250{,}000 - 90{,}000 = 160{,}000.$$

In the case of no mortgage, the living cost $C=100{,}000$ is computed by using $T=10$ and approx. 3\% of the house value as yearly rent. Thus,
$$ U_0 = 200{,}000 + 90{,}000 - 100{,}000 = 190{,}000,$$
where we assumed that only little less than $M_0$ is earned on average over the 10 years.
In contrast to the college admission, $U_0$ does not vanish and is even larger than $U_{01}$. It is therefore better not to get the mortgage when the likelihood of default is substantial. 

We consider an equal acceptance rate between both groups,  $P_0(\hat Y=1) = P_1(\hat Y=1)=0.8$. Conditionally on acceptance we assume that 95\% of the standard group does not default while 90\% of the protected group does not default. In this case the setting will clearly be unfair - and our question will be how to remedy this. First we compute the utility difference 
according to Equation \eqref{eq:EsU2}, 
\begin{align*}
\tilde \UD &= \begin{pmatrix}
     150{,}000\\ -30{,}000 \end{pmatrix} \cdot 
     \begin{pmatrix}
     0.76 - 0.72\\ 0.8-0.8\end{pmatrix} = 6{,}000.
\end{align*}
First, we observe that $\tilde U_2$ is negative (in contrast to being zero as in the college admission). 
Second, the small change in the default probability results in a difference which is 4\% of 150.000. Since default
probabilities can not be changed in this situation (although this is of course an important aspect of fairness)
we aim at changing the acceptance rate for the protected group to obtain a situation where both groups have equal utilities.

Quite surprising, decreasing the acceptance rate improves the utility for the protected group: we need to solve
$$ -30{,}000 \cdot (0.8 - p) = -60{,}000, $$
and we find $p=0.6$. Thus, with a reduced acceptance rate for the protected group of 0.6 the setting turns out to be fair in the sense that there is no disadvantage for the protected group.
The reason for this is that in the case of default the situation is disadvantageous. Thus, simply aiming at equal acceptance rates actually turns out to lead to a disadvantage in utility for the protected group - clearly an unwanted effect.

\section{Accounting for uncertainty}
Up to now we assumed that the probability measures were given and exactly known. This is of course not true in practice where probabilities have to be estimated. 
The estimated probabilities are subject to sampling uncertainty, and this uncertainty can be quantified by confidence intervals, see for example \cite{besse2018confidence}. In this section, we extend our previous setting by working with uncertainty sets, which could be obtained from confidence intervals or, in the spirit of Knightian uncertainty, by a combination with expert opinions or other data sources.

To simplify the approach we remain in the reduced setting of Equation \eqref{eq:EsU2}, while the standard setting can be treated in the same manner. 

To this end we denote the estimated probabilities by $\pi$ - the uncertainty interval will be represented by $ [\underline{\pi} ,  \overline{\mathrlap{\pi}\phantom{\underline{\pi}}}].$ In this spirit, the uncertainty interval for $\tilde \bP_s$ is denoted by the two intervals
\begin{align}
\begin{matrix}
        [ \underline{\pi}_s(Y=1, \hat Y=1) \, , \, 
        \overline{\mathrlap{\pi}\phantom{\underline{\pi}}}_s(Y=1, \hat Y=1) ], \quad \text{ and}\\[2mm]
        [\underline{\pi}_s(\hat Y = 1)
     \, , \,\overline{\mathrlap{\pi}\phantom{\underline{\pi}}}_s(\hat Y = 1)]
\end{matrix} 
\end{align}

\begin{proposition}
    Assume that $\tilde \bU \ge 0$. Then, the upper bound of the utility difference is given by
    \begin{align}
        \overline \UD = \tilde \bU \cdot \Pi^*
    \end{align}
    with 
    \begin{align}
        \Pi^* = \begin{pmatrix}
        \overline{\mathrlap{\pi}\phantom{\underline{\pi}}}_0(Y=1, \hat Y=1) - \underline{\pi}_1(Y=1, \hat Y=1) \\
        \overline{\mathrlap{\pi}\phantom{\underline{\pi}}}_0(\hat Y = 1) - \underline{\pi}_1(\hat Y = 1)
        \end{pmatrix}.
    \end{align}
\end{proposition}

\subsection{Confidence intervals}
Depending on the chosen approach, estimating confidence intervals could be challenging. While for logistic regression, a common statistical approach to binary classification, this is well-known, it might be less known that 
confidence intervals can also be constructed for machine-learning algorithms such as random forests. For example, one may rely on the jackknife estimator, see \cite{wager2014confidence}.

Several subtleties should, however, be taken into account: first, recall that we considered no covariables (up to now), i.e.\ the individuals which we consider should have the same or at least similar values of the  covariables. This could be the same credit rating, the same post code, a similar GPA, etc. Moreover, the data should be i.i.d.; this assumption may fail when observations span many years and should then also be taken into account.  Since smaller sample sizes lead to wider  confidence intervals, we expect that this issue is important in many applications.

\section{Outlook on a general setting}

In the general setting we introduce a probabilistic model of the default along the line of the famous Merton model, \cite{merton1974pricing}, with an adaptation to the mortgage case. A more detailed modelling would require treating imperfect information as in \cite{stiglitz1981credit,freyschmidt2009} and will be treated in future work. The following steps also serve as a guideline on how explicit biases in existing data could be modelled or how the approach could be applied in more complex questions which can not be treated by the standard setting in Equation \eqref{eq:EsU}.

 Let us assume that the future wealth $X_T$ is normally distributed with mean $\mu$ and standard deviation $\sigma$. Denote  $\Delta X_T = X_T - x_0$. At time $T$ we have that the utility of the borrower is given by
\begin{align*}
 U = \begin{cases}
 P_T + \Delta X_T - M_T & X_T > M_T, \\
 P_T\cdot (1-\lambda) + \Delta X_T - M_T & D_1, \\
 -x_0 & D_2,
 \end{cases}
 \end{align*}
 where $D_1$ denotes the default case where $P_T\cdot (1-\lambda) + X_T \ge  M_T$ and $D_2$ denotes the default case where $P_T\cdot (1-\lambda) + X_T <  M_T$.

\begin{figure}[t!]
    \centering
    \includegraphics[width=8cm]{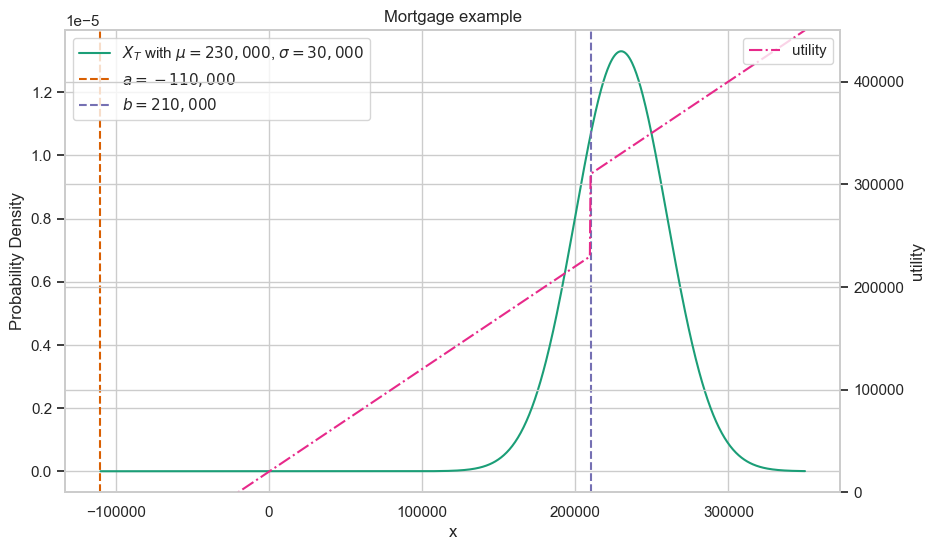}
    \caption{Illustration of the general setting - the future wealth $X_T$ is normally distributed and we plot the density, together with the two utility barriers - $D_1$ is between $a$ and $b$ - and the utility $U$.}
    \label{fig:mortgage2}
\end{figure}

As a first result we compute the distribution of $U$ together with its expected utility. Note that the case $11$ corresponds to no default, while $01$ corresponds to the default case (which in our case contains $D_1$ and $D_2$). 
Thus $P(Y=1|\hat Y=1) = P(X_T > M_T)$ while $P(Y=0|\hat Y=1) = 1-P(X_T> M_T) = P(D_1) + P(D_2)$. 

And, as a consequence,
\[
	P(Y=1,\hat Y=1)
	=P(\hat Y=1)\,
 \Phi\left(\frac{\mu-M_T}{\sigma}\right).
\]

Utilities in the case where the mortgage is not admitted is easily computed and equals $U_0 = x_0 + X_T$, such that $E[U|\hat  Y=0]= x_0 + \mu - C$ where $C$ includes additional costs due to lending. By $D=D_1 \cup  D_2$ we denote the event that default occurs, such that the probability of default equals $P(D)$. 
 \begin{lemma}
    Under the above assumptions, we have 
    \begin{align*}
        &P(X_T > M_T)= \Phi\Big(\frac{\mu-M_T}{\sigma}\Big) = 1-P(D), \\[2mm]
        &P(D_1)= \Phi\Big( \frac{M_T-\mu}{\sigma} \Big) - \Phi\Big( \frac{M_T-P_T\cdot(1-\lambda)-\mu}{\sigma} \Big), \\[2mm]
        &P(D_2)=\Phi\Big( \frac{M_T-P_T\cdot(1-\lambda)-\mu}{\sigma} \Big), \\[2mm]
        &E[U\ind{Y=1,\hat Y=1}] = \big( P_T + \mu - x_0 - M_T) \cdot \Phi\big( \frac{\mu - M_T}{\sigma}\Big) \\
        & \phantom{E[U_{11}]}+ \sigma \phi\Big( \frac{\mu - M_T}{\sigma}\Big), \\[2mm]
        &E[U\ind{Y=0,\hat Y=1}] = -x_0 \cdot P(D) + \big( P_T(1-\lambda) + \mu- M_T\big) \cdot P(D_1) \\
   & \phantom{E[U_{01}] }+ \sigma \Big( \phi\Big(\frac{M_T-P_T(1-\lambda)-\mu}{\sigma}\Big)
   - \phi\Big(\frac{M_T-\mu}{\sigma}\Big) \Big)  ,\\[2mm]
        &E[U | \hat Y =0]= x_0 + E[X_T] = x_0 + \mu - C.
    \end{align*}
 \end{lemma}

For an illustration, we compute the related values in correspondence to our previous mortgage example. We then have $ M_T=M_0=200{,}000$, $P_T=400{,}000$, $x_0 = 90{,}000$ and $C=100{,}000$. For illustration we expect that the average future wealth is higher than the mortgage, such that $\mu = M_T +20{,}000$ (not too high to see a significant default probability) and $\sigma = 30{,}000$ together with a $\lambda=0.2$, see Figure \ref{fig:mortgage2}. The default probability then computes to 25\% and $E[U\ind{Y=1,\hat Y=1}]=256{,}000$, $E[U\ind{Y=0,\hat Y=1}]=54{,}000$ and $E[U\ind{\hat Y=0}]=220{,}000 \cdot P(\hat Y=0) $. This illustrates that, as in the university admission, there is a possibly severe monetary loss in the case when the mortgage is admitted, $\hat Y=1$ and cannot be paid back, $Y=0$. This  illustrates that the utility consequences of granting a mortgage can differ substantially between the default and non-default cases. They also show that these consequences cannot be captured by acceptance and default probabilities alone and should be taken into account in fairness considerations relating to mortgages.

Analysing fairness in this context therefore requires estimating the above quantities for the considered groups. The computation of utility differences under several model assumptions is then straightforward with the results obtained above. A detailed  discussion, however, is beyond the scope of this short paper.

Summarizing, this small illustration of a general utility-based approach confirms the findings in our reduced-setting: fairness measured only on statistical measures is out of context and may be highly unfair if measured in utilities. To mitigate unfairness in the mortgage example not only acceptance rates have to be adjusted but also precautionary measures have to be taken into account to lower the highly unwanted effects when default occurs.

 \section{Conclusion}

This paper introduces a simple and tractable framework for measuring fairness in terms of utilities. We show that existing fairness measures do not necessarily guarantee fair outcomes from this utility-based perspective. In particular, the concept of $\varepsilon$-fairness may lead to misleading assessments of fairness. We also provide illustrative examples and mathematical tools that may serve as a useful basis for further applications.

\section{Acknowledgments}
We want to thank Alaa Awad, Daniel Feuerstack, Wilfried Hinsch, Lars Niemann and Jake Robertson for their discussions and support on the subject and the initiative Adaptive Governance of Emerging Technologies (AdGovEm) for its support.


\section{Disclosure of Interests}
The authors have no competing interests to declare that are relevant to the content of this article.

\bibliographystyle{splncs04}

\newpage

\begin{appendix}

\section{Proofs}
\setcounter{proposition}{0}
\begin{proposition} 
Assume that there are only two groups, i.e.\ $N=1$. Then, 
    the expected utility  is given by
    \begin{align*}
        E_s[U] &=  P_s(Y=1,\,  \hat Y=1)  \cdot ( U_{11} - U_{01} ) \notag \\
        &+P_s(Y=0, \, \hat Y=0) \cdot( U_{00} - U_{10}) \notag \\
        & +P_s(\hat Y=1) \cdot (U_{01}-U_{10}) +
     U_{10}.
    \end{align*}
\end{proposition}

\begin{proof}
The overall utility is given by summing over all possible states, i.e.
	$$ U = \sum_{k,j=0}^1 \ind{Y = k, \hat Y = j} U_{kj}. $$
Hence, the expected utility equals 
	\begin{align} \label{EUi:1}
		E_s[U] &= \sum_{k,j=0}^1 P_s(Y = k, \hat Y = j) \cdot U_{kj} . 
	\end{align}
	Now let us concentrate on $j=1$. We obtain that
	$$ P_s(Y = 0 \, , \, \hat Y = 1) = P_s( \hat Y=1) -  P_s(Y=1 \, , \,  \hat Y=1). $$
	Hence, inserting this into Equation \eqref{EUi:1}, 
\begin{align}\label{eq:3.4}
		E_s[U \ind{\hat Y =1 }] &= P_s(\hat Y=1) \cdot U_{01} \notag \\
  &+ P_s(Y=1 \,,\, \hat Y=1) \cdot ( U_{11} - U_{01} ). 
	\end{align}
	Similarly, $P_s(Y = 1 \,,\, \hat Y = 0) = P_s(\hat Y=0)- P_s(Y = 0 \,,\, \hat Y = 0)$, such that
	\begin{align}\label{eq:3.5}
		E_s[U \ind{\hat Y = 0}] &= P_s(\hat Y=0) \cdot U_{10} \notag \\
  &+ P_s(Y=0 \,,\, \hat Y=0) \cdot( U_{00} - U_{10})
	\end{align}
	and the conclusion follows by summing \eqref{eq:3.4} and \eqref{eq:3.5} and using that $P_s(\hat Y=0) = 1- P_s(\hat Y=1)$.
\end{proof}

\setcounter{proposition}{2}
\begin{proposition}
Assume that \eqref{eq:d2} holds.
	 Then, there exist for every $K \ge 0$, there exist parameters $U_{00},\dots,U_{11}$ such that
	$$ \UD \ge K. $$
\end{proposition}
\begin{proof}
    The core of the argument is representation \eqref{eq:UD with PD}, which we repeat for convenience,
    \begin{align*}
        \UD = \begin{pmatrix}
     U_{11}-U_{01}\\ U_{00}-U_{10} \\  U_{01}-U_{10}\end{pmatrix} \cdot 
     \begin{pmatrix}
        \PD(Y=1, \hat Y=1) \\
        \PD(Y=0, \hat Y=0) \\
        \PD(\hat Y = 1).
    \end{pmatrix}
    \end{align*}
We set $U_{01}=U_{10}$, such that $\UD_3=0$.  Equation \eqref{eq:d2} implies a number of cases which we can treat similarly. Consider for example  $\PD(Y=1,\,  \hat Y=1) > 0$.
We choose $U_{00}=U_{10}$ to obtain $\UD_2=0$. Then, we choose $U_{10}=0$ and $U_{11}=K \cdot \PD(Y=1,\,  \hat Y=1)^{-1} >0$, such that 
	\begin{align}
		\UD = ( U_{11} - U_{01} ) \cdot \PD(Y=1,\,  \hat Y=1)  = K.
	\end{align}
	The other results follow in a similar way.
\end{proof}

\setcounter{proposition}{4}
\begin{proposition}
    Assume that base-rate parity and  equalized odds hold and that $U_{01} = U_{10}$. Then, there is no disadvantage.
\end{proposition}
\begin{proof}
We begin by noting that Bayes' formula implies that
\begin{align*}
    P_s(Y=i) \cdot P_s(\hat Y=i|Y=i) = P_s(\hat Y=i , Y=i).
\end{align*}
Thus, statistical parity together with equalized odds imply equal joint distributions on the diagonal. Since $U_{01}=U_{10}$, we obtain that $U_3=0$ and the claim follows from Equation \eqref{eq:UD with PD}.
\end{proof}

\begin{proposition}Assume that equality of the conditional use accuracy holds. 
\begin{enumerate}[(i)]
    \item If $P_0(\hat Y=1)=P_1(\hat Y=1)$, then there is no disadvantage.
    \item If  $|P_0(\hat Y=1)=P_1(\hat Y=1)|\le \varepsilon$ and the utility differences are bounded by $K>0$, then there is no disadvantage larger than
    $$ (1+2 \Gamma) \cdot  \varepsilon \cdot K.$$
\end{enumerate}
\end{proposition}
\begin{proof}
For the first claim, observe that 
\begin{align*}
    \PD(Y=1, \hat Y=1) &= P_0(Y=1,\hat Y=1) - P_1(\hat Y=1).
\end{align*}
Note that by Bayes' theorem,
$$ P_s(Y=1,\hat Y=1) = P_s(\hat Y=1) \cdot P_s(Y=1 |\hat Y=1)$$
for $s=0,1$. Since CUA implies that the conditional probabilities $P_s(Y=1 |\hat Y=1)$, $s=0,1$ are equal it remains to consider the probabilities $P_s(\hat Y=1). $ This is exactly our assumption in (i) and it follows that the probability difference vanishes, i.e.
$$
\PD(Y=1, \hat Y=1) =0.
$$
For the probability difference $\PD(Y=0, \hat Y=0)$ observe that the assumption in (i) also implies $P_0(\hat Y=0) = P_1(\hat Y=0)$ and we obtain in a similar manner
$$ 
\PD(Y=0, \hat Y=0) =0.
$$
Since we also have $\PD(\hat Y=1)=0$, Proposition \ref{prop: 3.6} implies absence of any disadvantage. 

For the second claim we estimate the utility difference using Equation \eqref{eq:EUi 1} by
$$ 
|\UD|  \le K \cdot \parallel \PD \parallel_\infty 
$$
where $\parallel \PD \parallel_\infty $ denotes the maximum of the absolute values of the rows of the vector $\PD$.
We examine the rows of $\PD$ separately. For the first row we argue as above, using Bayes' theorem, and obtain that (again using CUA)
\begin{align*} 
|\PD(Y=1,\hat Y=1)| &\le P_0(Y=1 | \hat Y=1) \cdot |\PD(\hat Y=1)|  \\
& \le  P_0(Y=1 | \hat Y=1) \cdot \varepsilon \le \Gamma \cdot \varepsilon. \end{align*}
In an analogous manner we obtain a similar estimate for the second row and $\varepsilon \cdot K$ as a bound for the third row and the proof is finished.
\end{proof}

\begin{proposition}
Independence of $(Y, \hat Y)$ and $S$ implies absence of a disadvantage. 
\end{proposition}
\begin{proof}
Note that independence implies for all $i,j$ that 
	\begin{align*}
		P_0(Y=i \,, \,\hat Y=j ) &= 
		P(Y=i \,, \,\hat Y=j \, | \, S=0) \\
  &= P(Y=i \,, \,\hat Y=j) \\
  &= P_1(Y=i \,, \,\hat Y=j ).
	\end{align*}
Now, Proposition \ref{prop: 3.6} yields the result.
\end{proof}

\setcounter{lemma}{0}
\begin{lemma}
The expected utility in the reduced setting computes to
\begin{align*}
    E_s[\tilde U] = \tilde \bU \cdot \tilde \bP_s + U_0
\end{align*}
for $s \in \{0,1\}$ with
\begin{align*}
    \tilde \bU := \begin{pmatrix}
     U_{11}-U_{01}\\ U_{01}-U_{0}\end{pmatrix}, \qquad 
    \tilde \bP_s := \begin{pmatrix}
        P_s(Y=1, \hat Y=1) \\
        P_s(\hat Y = 1).
    \end{pmatrix}
\end{align*}
\end{lemma}

\begin{proof}
As in Proposition 1, 
\begin{align*}
E_s[U] &= U_{11} \cdot P_s(Y=1, \hat Y= 1) + U_{01} \cdot P_s(Y=0, \hat Y=1) \\
& + U_0 P_s(\hat Y= 0).
\end{align*}
Since $P_s(\hat Y= 0) = 1- P_s(\hat Y=1)$ and 
\begin{align*}
P_s(Y=0,\hat Y=1) + P_s(Y=1, \hat Y=1) = P_s(\hat Y=1),
\end{align*}
we obtain that
\begin{align*}
E_s[U] &= (U_{11}-U_{01}) \cdot P_s(Y=1, \hat Y=1) \\ 
	&+ (U_{01}-U_0) \cdot P_s(\hat Y=1) + U_0.
\end{align*}
Putting this into the representation of the lemma, the result is proven.
\end{proof}

 \begin{lemma}
 	Under the above assumptions, we have 
 	\begin{align*}
 		&P(X_T > M_T)= \Phi\Big(\frac{\mu-M_T}{\sigma}\Big) = 1-P(D), \\[2mm]
 		&P(D_1)= \Phi\Big( \frac{M_T-\mu}{\sigma} \Big) - \Phi\Big( \frac{M_T-P_T\cdot(1-\lambda)-\mu}{\sigma} \Big), \\[2mm]
 		&P(D_2)=\Phi\Big( \frac{M_T-P_T\cdot(1-\lambda)-\mu}{\sigma} \Big), \\[2mm]
 		&E[U\ind{Y=1,\hat Y=1}] = \big( P_T + \mu - x_0 - M_T) \cdot \Phi\big( \frac{\mu - M_T}{\sigma}\Big) \\
 		& \phantom{E[U_{11}]}+ \sigma \phi\Big( \frac{\mu - M_T}{\sigma}\Big), \\[2mm]
 		&E[U\ind{Y=0,\hat Y=1}] = -x_0 \cdot P(D) + \big( P_T(1-\lambda) + \mu - M_T\big) \cdot P(D_1) \\
   & \phantom{E[U_{01}] }+ \sigma \Big( \phi\Big(\frac{M_T-P_T(1-\lambda)-\mu}{\sigma}\Big)
   - \phi\Big(\frac{M_T-\mu}{\sigma}\Big) \Big)  ,\\[2mm]
        &E[U|\hat Y =0]= x_0 + E[X_T] = x_0 + \mu - C.
 	\end{align*}
 \end{lemma}
 \begin{proof}
Since $X_T$ is normally distributed,  $X_T=M_T + \sigma \xi$ with mean $M_T$ and standard deviation $\sigma$. This already gives the first result. Next, $P(D_1) = P(M_T-P_T(1-\lambda) < X_T \le M_T)$ which implies the second and the third equation.
The utility in the case $Y=1,\hat Y=1$ is given by
\begin{align*}
E\Big[ \big(P_T + X_T - x_0 - M_T \big)&\ind{X_T > M} \Big] =\big( P_T  +\mu - x_0 - M_T \big) \cdot P(X_T>M) \\
    &+ \sigma \cdot E \Big[ \xi \, \ind{\xi > \frac{M_T-\mu}{\sigma}} \Big].
\end{align*}
Noting that 
\begin{align*}
\int_C^\infty x \frac 1 {\sqrt{2 \pi}}e^{-\frac{x^2}2} = \frac 1 {\sqrt{2 \pi}}e^{-\frac{C^2}2} = \phi(C),
\end{align*}
the fourth equation follows. In a similar way,
\begin{align*}
    E\Big[ \big(P_T(1-\lambda) + &X_T - x_0 - M_T \big) \cdot \ind{M_T-P_T(1-\lambda)<X_T \le M_T} \Big] \\
    &+ E\Big[ - x_0 \cdot\ind{X_T \le M_T-P_T(1-\lambda)} \Big] \\[2mm]
    &= -x_0 \cdot P(D_1+D_2) +\big(P_T(1-\lambda) + \mu  - M_T \big) \cdot P(D_1) \\[2mm]
    & + \sigma \cdot E \Big[ \xi \, \ind{M_T-P_T(1-\lambda)<X_T \le M_T} \Big].
\end{align*} 
Noting that 
$
\int_{C_1}^{C_2} x \frac 1 {\sqrt{2 \pi}}e^{-\frac{x^2}2} =\phi(C_1)-\phi(C_2),
$
the fifth equation follows. Since the last equation is immediate, the proof is finished.
 \end{proof}

\end{appendix}
 
 \end{document}